\documentclass[journal]{IEEEtran}
\pdfoutput=1 

\usepackage{amsmath}
\ifCLASSINFOpdf
\else
   \usepackage[dvips]{graphicx}
\fi
\usepackage{url}

\usepackage{graphicx}
\usepackage{cite}
\usepackage{xcolor}

\begin{document}

\title{Object Remover Performance Evaluation Methods \\ using Class-wise Object Removal Images}

\author{Changsuk Oh, Dongseok Shim, Taekbeom Lee, and H. Jin Kim 
\thanks{Changsuk Oh, Taekbeom Lee, and H. Jin Kim are with Aerospace Engineering, Seoul National University, Seoul 08826, South Korea (e-mail: santgo@snu.ac.kr, ltb1128@snu.ac.kr, hjinkim@snu.ac.kr).}
\thanks{Dongseok Shim is with Artificial Intelligence, Seoul National University, Seoul 08826, South Korea (e-mail: tlaehdtjr01@snu.ac.kr).}

}

\markboth{}
{Shell \MakeLowercase{\textit{et al.}}: Bare Demo of IEEEtran.cls for IEEE Journals}
\maketitle

\begin{abstract}
Object removal refers to the process of erasing designated objects from an image while preserving the overall appearance, and it is one area where image inpainting is widely used in real-world applications. The performance of an object remover is quantitatively evaluated by measuring the quality of object removal results, similar to how the performance of an image inpainter is gauged. Current works reporting quantitative performance evaluations utilize original images as references. In this letter, to validate the current evaluation methods cannot properly evaluate the performance of an object remover, we create a dataset with object removal ground truth and compare the evaluations made by the current methods using original images to those utilizing object removal ground truth images. The disparities between two evaluation sets validate that the current methods are not suitable for measuring the performance of an object remover. Additionally, we propose new evaluation methods tailored to gauge the performance of an object remover. The proposed methods evaluate the performance through class-wise object removal results and utilize images without the target class objects as a comparison set. We confirm that the proposed methods can make judgments consistent with human evaluators in the COCO dataset, and that they can produce measurements aligning with those using object removal ground truth in the self-acquired dataset. 

\end{abstract}


\IEEEpeerreviewmaketitle

\section{Introduction}
\IEEEPARstart{O}{bject} removal is one area where image inpainting is extensively used in real-world applications. Unlike assessing the quality of inpainted images, gauging the quality of object removal results poses considerable challenges. This arises from the fact that an original image and an object removal result differ in the presence of the removal target. Consequently, most previous works \cite{liu2018image, yi2020contextual, cao2021learning, rombach2022high, zeng2020high, suvorov2022resolution, zeng2022aggregated, 9195134} reporting object removal results present only qualitative results. Only a few, such as \cite{angah2020removal,kottler20223gan,9446636}, quantitatively evaluate the quality using full-reference (FR) methods which employ original images as references. In this letter, we demonstrate that the FR methods used in the previous works are not suitable for evaluating object removal results and introduce novel FR methods designed for evaluating the performance of an object remover.

\begin{figure}[t]
  \centering
  \includegraphics[width=0.4\textwidth]{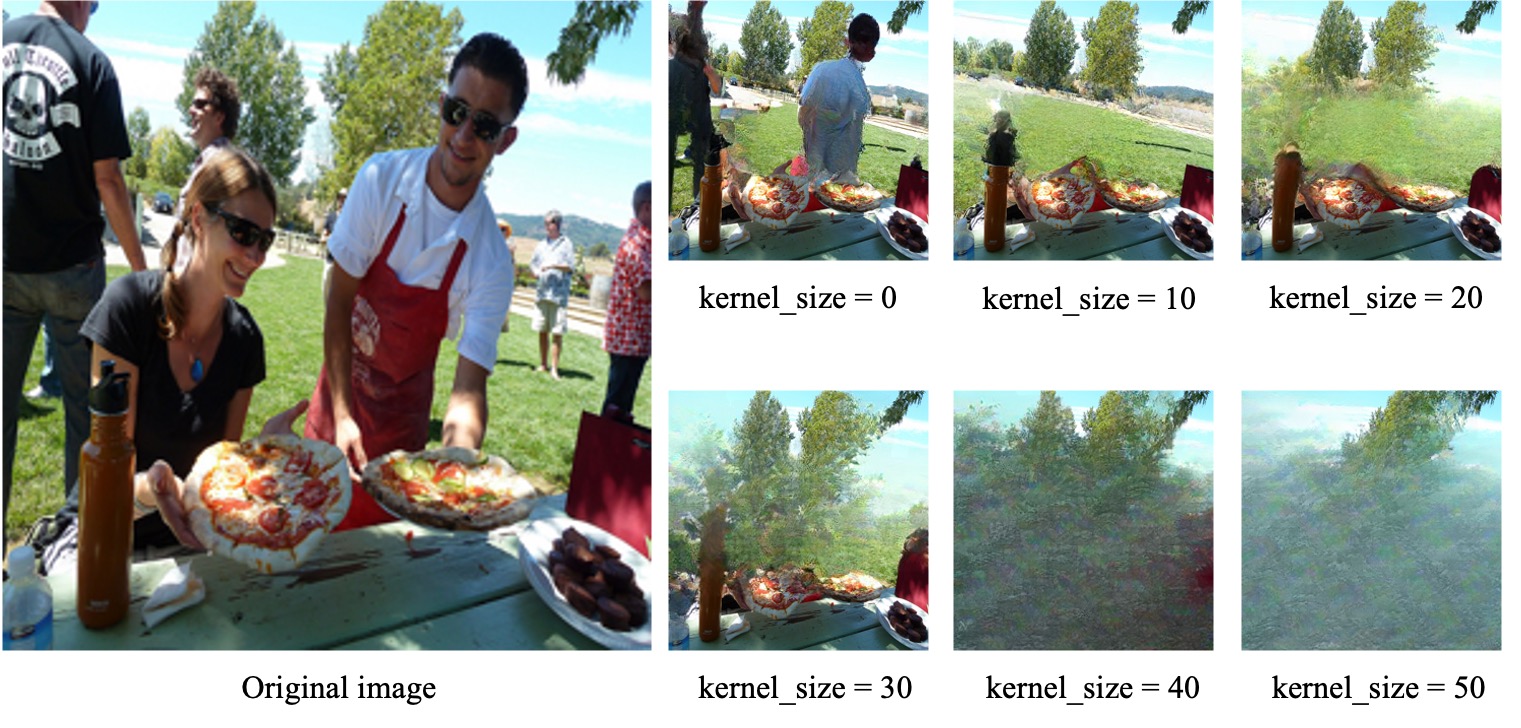}
\vspace{-3mm}
  \caption{Object removal results on the COCO dataset. We use Lama \cite{suvorov2022resolution} for removal and \textit{human} class objects are removed. Previous evaluation methods judge that the object remover using $kernel\_size=0$ performs the best, while our methods evaluate that the object remover using $kernel\_size=10$ erased the removal target in the most plausible way.}
    \label{fig:qualitative}
\vspace{-6mm}
\end{figure}

\begin{figure*}[ht!]
  \centering
  \includegraphics[width=0.7\textwidth]{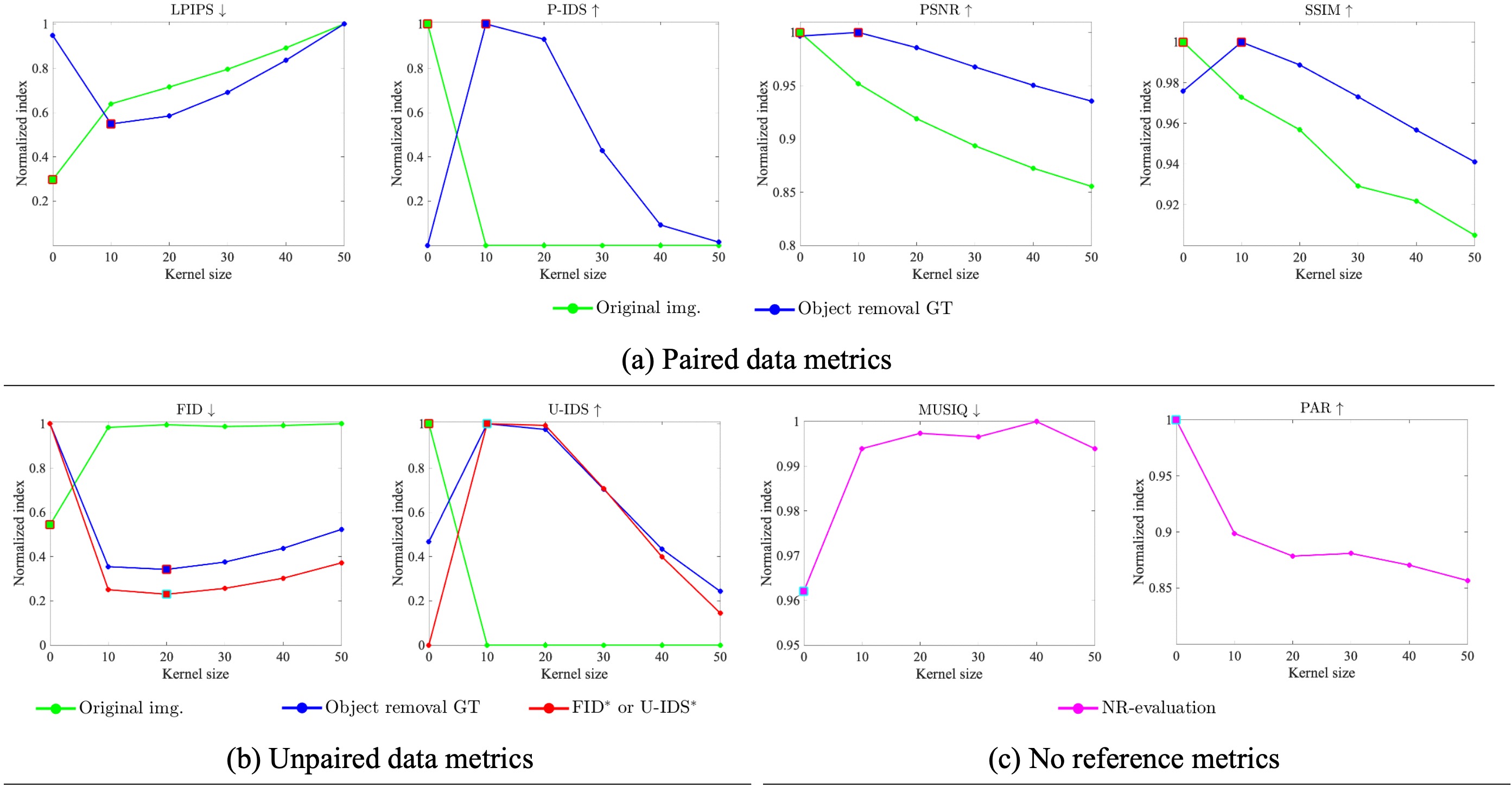}
  \vspace{-3mm}
  \caption{Object remover performance evaluations on the proposed CARLA dataset. We use Lama \cite{suvorov2022resolution} for removal. The score of each method is divided by the highest score of each method. $\uparrow$ and $\downarrow$ indicate higher is better and lower is better, respectively. The object remover which is judged to have the best performance for each evaluation method is indicated with a rectangle. }
    \label{fig:evaluation_metric}
    \vspace{-6mm}
\end{figure*}

FR methods can be categorized into two groups. LPIPS \cite{zhang2018unreasonable}, PSNR \cite{wang2004image}, SSIM \cite{wang2004image}, and P-IDS \cite{zhao2021comodgan} require a pair of original and completed images, while FID \cite{heusel2017gans} and U-IDS \cite{zhao2021comodgan} can be measured using unpaired data. To validate whether the paired or unpaired data methods can properly evaluate the performance with  original images, we generate a dataset with object removal ground truth (GT) using virtual environments. Then, we compare the evaluations made by the FR methods using the original images with those utilizing the object removal GT images. As shown in Fig. \ref{fig:evaluation_metric}, we confirm the disparities between the two evaluation sets, which indicates that the FR methods using original images as references cannot properly measure the quality of object removal results. We report details about the experiment in Section \ref{section:existing}.

We propose new unpaired methods to properly judge which object remover performs better in the absence of object removal GT. The proposed evaluation methods assess the performance of an object remover using the class-wise object removal tasks, where all objects in an image belonging to a target class are designated as removal targets. The proposed methods utilize images that do not contain an object of a target class for the comparison. In this scenario, both activation vectors obtained from object removal results and those obtained from the comparison set do not have features from target class objects. Therefore, unpaired methods can focus on evaluating whether the object removal results are as realistic as the comparison set. We demonstrate that the proposed methods can produce evaluations consistent with those made by humans in the COCO dataset, and that they can generate measurements aligning with those using object removal GT in the self-acquired virtual environment dataset. The proposed framework has the following contributions:
\begin{itemize}
    \item We experimentally validate that the current object remover evaluation methods, which utilize original images as references, cannot properly measure the performance. To compare the evaluations made by the FR methods using original images to those utilizing object removal ground truth images, we generate a dataset with object removal ground truth images. 
    
    \item The proposed methods measure the performance of an object remover using class-wise object removal results and the comparison set composed of images without target class objects. The proposed methods can evaluate the performance of an object remover without generating object removal ground truth.
    
    \item Experiments on the images from various environments demonstrate that the proposed methods can properly evaluate the object remover performance regardless of the input images’ style. This suggests that the proposed methods can be applied to model selection during object remover training or performance assessment of off-the-shelf object removers, which enables the training or selecting models that produce superior object removal results for query images from various environments.
\end{itemize}
\section{Object Remover Performance Evaluation} \label{section:2}

\subsection{Current object remover performance evaluation methods}
\label{section:existing}
\cite{angah2020removal,kottler20223gan,9446636} measure the performance of an object remover using the quality of the object removal results generated by the object remover, similar to how the performance of an image inpainter is gauged. As the large-scale image datasets \cite{zhou2017places,karras2018progressive,lin2014microsoft}, widely used for training image inpainting models, do not contain the object removal GT images, \cite{angah2020removal,kottler20223gan,9446636} utilize original images as a comparison set. We infer that the FR method utilizing original images as references cannot properly evaluate the quality because an original image and object removal result differ in the presence of removal targets. To experimentally validate this, we generate a dataset with object removal GT. Then, we check whether the evaluations using the object removal GT are consistent with those using original images. Additionally, we also examine whether a non-reference (NR) inpainted image quality assessment (PAR \cite{zhang2022perceptual}) and a NR image quality assessment (MUSIQ \cite{ke2021musiq}) methods can make judgment consistent with evaluations using object removal GT images.  

\begin{figure}[t!]
  \centering
  \includegraphics[width=0.3\textwidth]{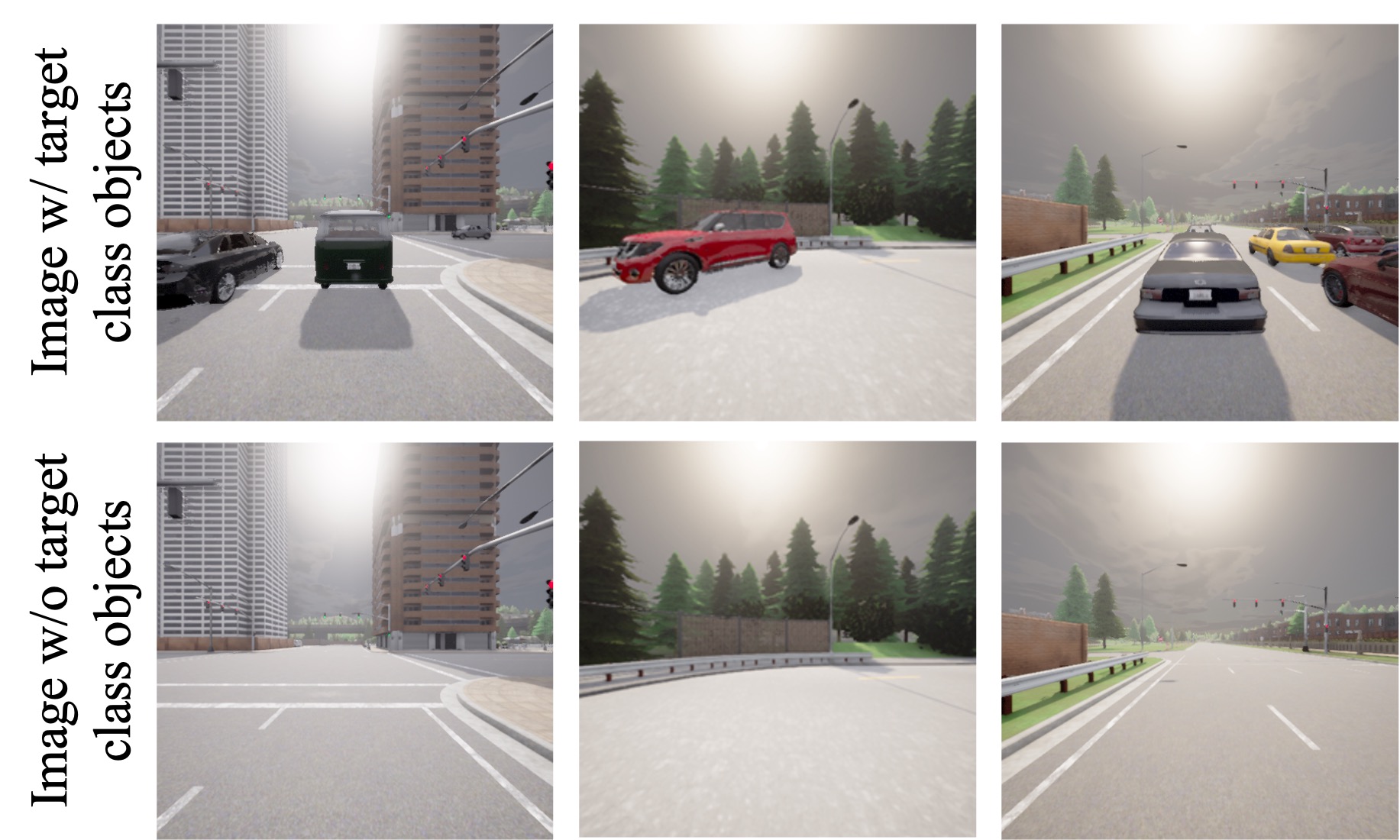}
  \caption{Image samples of the CARLA dataset with object removal ground truth. }
    \label{fig:dataset}
\vspace{-6mm}
\end{figure}

\textbf{Implementation details. }
We utilize five maps provided by the CARLA simulator to acquire images from complex environments. For each map, we capture 5K images twice, resulting in a total of 50K images. The images of the first set contain vehicles in the scene, while the images of the second set are captured in the same scenes, but without surrounding vehicles. We only acquire an image when the ego-vehicle is moving at a certain speed or higher to avoid capturing the same scene multiple times. We use the semantic segmentation maps provided by the CARLA simulator to designate the removal targets, and instances belonging to the car class are set as the removal target. Fig. \ref{fig:dataset} shows images of the dataset. We only use images in the first set whose masks cover more than 1\% of the images for the object removal experiment. We build six object removers using an inpainting model and six different types of masks. Using the GT segmentation maps, we obtain a set of masks that pixels from the target class objects are marked. Then, we create five variant mask sets by manipulating the original mask set using the $opencv.dilate(kernel\_size)$ function. 

\textbf{Evaluation results. }
Evaluations of six object removal sets are presented in Fig. \ref{fig:evaluation_metric}. Both paired and unpaired data methods yield different judgments when using object removal ground truth and when using the original images. Specifically, in case of using original images, all evaluation methods judge that the object remover using $kernel\_size = 0$ has the highest performance. However, when using object removal ground truth, PSNR, SSIM, LPIPS, P-IDS, and U-IDS evaluate that removal results generated by the object remover using $kernel\_size = 10$ have the highest quality, and FID judges that the object remover using $kernel\_size = 20$ can best erase removal targets. The results indicate that the paired and unpaired data methods using original images as references are not suitable as object removal evaluation methods in the absence of object removal ground truth.

For the paired data metrics, as they evaluate how similar a completed image is to the original image, the estimated quality of the completed image is inversely proportional to the size of the input mask, as shown in Fig. \ref{fig:evaluation_metric}(a). We also observe the inversely proportional tendency when using the unpaired methods (FID and U-IDS), as shown in Fig. \ref{fig:evaluation_metric}(b). In the experiment, we remove all the vehicles in the images and employ the original images containing the vehicles as a comparison set. Both FID and U-IDS use pre-trained CNNs to convert images into activation vectors and evaluate the quality using them. The activation vectors obtained from images where removal targets are poorly removed contain more features related to the target class objects compared to those whose removal targets are well removed. As the existing unpaired data methods do not place any restrictions on the images of the comparison set, the features from the vehicle class objects are included in the activation vectors of the comparison set images. Consequently, even in the cases where the objects are poorly erased, the existing methods cannot lower the score of the completed images because the activation vectors of the comparison set have similar features. 

Two non-reference (NR) methods (MUSIQ and PAR) also judge that the quality decreases as the mask size increases, which is inconsistent with any evaluations using the object removal GT images. This result indicates that MUSIQ and PAR are also not appropriate to measure the quality of object removal results conducted on the images obtained from virtual environments.

\subsection{Proposed method}
\label{section:proposed}
We propose novel evaluation methods designed to assess the performance of an object remover. First, we search a task where it is possible to obtain a comparison set that works as object removal GT. Then, we evaluate the performance through the unpaired methods using the comparison set. Unlike other image generation tasks, object removal requires masks as inputs to specify removal targets. To obtain masks without manual process, our method utilizes images with semantic segmentation annotation as inputs. 

The proposed method explolits class-wise object removal results to evaluate the performance. Although obtaining class-wise object removal GT of public datasets is not feasible, it is possible to collect a sufficient number of images without target class objects. In other words, even if we cannot use object removal GT for the evaluation, we can utilize images without target class objects as a comparison set to calculate the unpaired methods.  This set is identical to the object removal GT of class-wise object removal results in terms of the absence of target class objects. Therefore, our method utilize class-wise object removal results as a query set and images without target class objects as a comparison set to calculate FID and U-IDS. 

\begin{figure*}[ht!]
  \centering
  \includegraphics[width=0.95\textwidth]{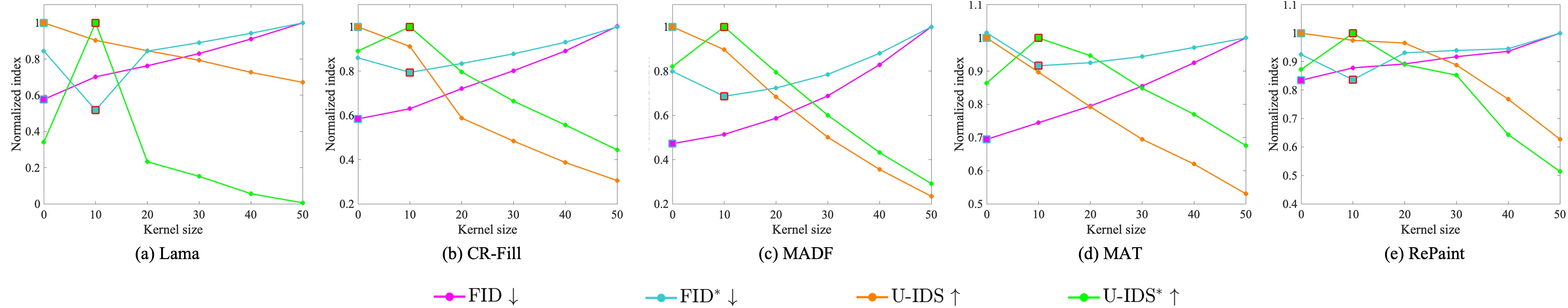}
\vspace{-3mm}
  \caption{Performance evaluations made by the unpaired data methods (FID and U-IDS) and proposed methods (FID$^*$ and U-IDS$^*$) on the COCO dataset. The score of each method is divided by the highest score of each method. $\uparrow$ and $\downarrow$ indicate higher is better and lower is better, respectively. The object remover which is judged to have the best performance for each evaluation method is indicated with a rectangle.}
    \label{fig:quantitative}
\vspace{-5mm}
\end{figure*}

In this scenario, the activation vectors of query images generated by a well-performing object remover do not contain features from target class objects. Then, both FID and U-IDS can focus on evaluating whether the query set is as realistic as the comparison set because the activation vectors acquired from the comparison set also do not contain features from target class objects. Conversely, if we generate a query set using a poorly performing object remover, activation vectors of the query images contain features of the target class objects, as the object removal results have more poorly-erased regions. The features of the target class objects make discrepancies between activation vector sets obtained from the query and comparison sets, which can lower the score of the poorly performing object remover.  

We refer to FID and U-IDS using class-wise object removal results and the comparison set without target class objects for the calculation as FID$^*$ and U-IDS$^*$, respectively. FID$^*$ is defined as follows:
\begin{align}
    \text{FID}^*&(X,X') =  \text{FID}(X_n, X') \notag \\
    & = \lVert \mu_n - \mu' \lVert_2^2 + tr (\Sigma_n + \Sigma' - 2(\Sigma_n \Sigma')^{1/2} ),   
\label{eq:fid_star}
\end{align} 
where $(\mu_n, \Sigma_n)$ and $(\mu'$,$\Sigma'$) are the mean and covariance matrix of the real data (without target class objects) distribution $X_n$ and class-wise object removal data distribution $X'$, respectively. $X$ indicates a real data distribution. 
We define U-IDS$^*$ as follows:
\begin{align}
    &\text{U-IDS}^*(X, X') = \text{U-IDS}(X_n, X') \notag \\
    &= \frac{1}{2}\Pr_{x_n \in X_n}{\left\{f(\mathcal{I}(x_n)) < 0\right\}} + \frac{1}{2}\Pr_{x' \in X'}{\left\{f(\mathcal{I}(x')) > 0\right\}}.
\end{align}
$\mathcal{I}$ is the pre-trained Inception v3 \cite{szegedy2016rethinking} that converts an input image to feature vector of 2048 dimensions. $f$ is a linear support vector machine that outputs a positive value when considering the input sample as real.

Fig. \ref{fig:evaluation_metric}(b) shows evaluations made by the proposed methods. We use the images of the second set whose counterpart is not used for the object removal experiments to generate a comparison set. FID$^*$ and U-IDS$^*$ evaluate that object remover using $kernel\_size = 20$ and $kernel\_size = 10$ can best erase removal targets, respectively, which are consistent with evaluations made by FID and U-IDS using object removal ground truth images. This implies that we can evaluate the performance of object remover using FID$^*$ and U-IDS$^*$ without object removal GT.

\section{Experiments}

We experiment whether it is also possible to properly evaluate the performance of an object remover using a public image dataset and our method. Similar to \cite{cao2021learning}, we use images whose masks cover 5--40\% of the images. As shown in the following experiment, our method needs the use of more than 9K query images for a reliable evaluation. Therefore, we use the COCO dataset and \textit{person} class as a target class. The COCO dataset contains over 12K images where the \textit{person} class objects occupy 5--40\% of the pixels, and it has more than 35K images without \textit{person} class objects. We use five image inpainters (Lama \cite{suvorov2022resolution}, CR-Fill \cite{zeng2021cr}, MADF \cite{zhu2021image}, MAT \cite{li2022mat}, and RePaint \cite{lugmayr2022repaint}) as task networks. Similar to the experiments in Section \ref{section:2}, we evaluate which model among the object removers, generated by a single task network and six masks sets, performs the best.

 As the COCO dataset does not have object removal GT, we compare the estimated judgments made by evaluation methods to human evaluations. We conduct an user study to collect human judgments on which object remover best erases removal targets while preserving the remaining parts of original images. We utilize Lama \cite{suvorov2022resolution} to generate object removal results for the human evaluations. 23 users are asked to choose one of the six object removal results that the target class objects are removed in the most visually plausible way. Each user evaluates 10 image sets, and the judgments are presented in Table \ref{table:votes}. Human evaluators gauge that the object remover using $kernel\_size=10$ erases removal targets in a most plausible way. Fig. \ref{fig:qualitative} shows an original image and six object removal results. 

\begin{table}[t!]

\centering
\caption{Human evaluations on the performance of object removers.}
\vspace{-1mm}
\begin{tabular}{c|cccccc}
\hline
Kernel size & 0 & 10 & 20 & 30 & 40 & 50 \\ \hline
Votes & 9 & 134 & 54 & 19 & 14 & 0 \\ \hline
\end{tabular}
\label{table:votes}
\vspace{-6mm}
\end{table}

\textbf{Results. } Fig. \ref{fig:quantitative}(a) shows the evaluation results made by the unpaired data methods on the object removal results generated by Lama \cite{suvorov2022resolution}. FID and U-IDS judge that the performance of the object remover using $kernel\_size=0$ is the best. On the other hand, FID$^*$ and U-IDS$^*$ assess that the performance of the object remover using $kernel\_size=10$, in line with human evaluations, is the best. Furthermore, we confirm identical tendencies in the results obtained using the other task networks (CR-Fill, MADF, MAT, and RePaint). Specifically, FID and U-IDS, regardless of the task network, conclude that the performance of the object remover using the smallest masks ($kernel\_size=0$) is the best, which is inconsistent with judgement made by human evaluators. However, FID$^*$ and U-IDS$^*$ judge that the object remover using $kernel\_size=10$ erases removal targets best, which is consistent with assessments made by human evaluators. This result demonstrates that our method can properly evaluate the performance of an object remover using a public dataset, regardless of the backbone network of the object remover.

\textbf{Input image style. } 
In real-world applications, object removal is performed on images of various styles. In Section \ref{section:2}, we check that NR methods, utilizing pre-trained networks trained on images acquired from real environments, are not suitable for assessing the quality of object removal results whose original images are acquired from virtual environments. On the other hand, the proposed methods successfully evaluate the quality of object removal results conducted on images obtained from both real and virtual environments. The proposed methods can properly evaluate the performance using images from both real and virtual environments because the unpaired data methods can assess the quality of a query set when the styles of a comparison set and query set are identical.

\begin{figure}[t]
  \centering
  \includegraphics[width=0.45\textwidth]{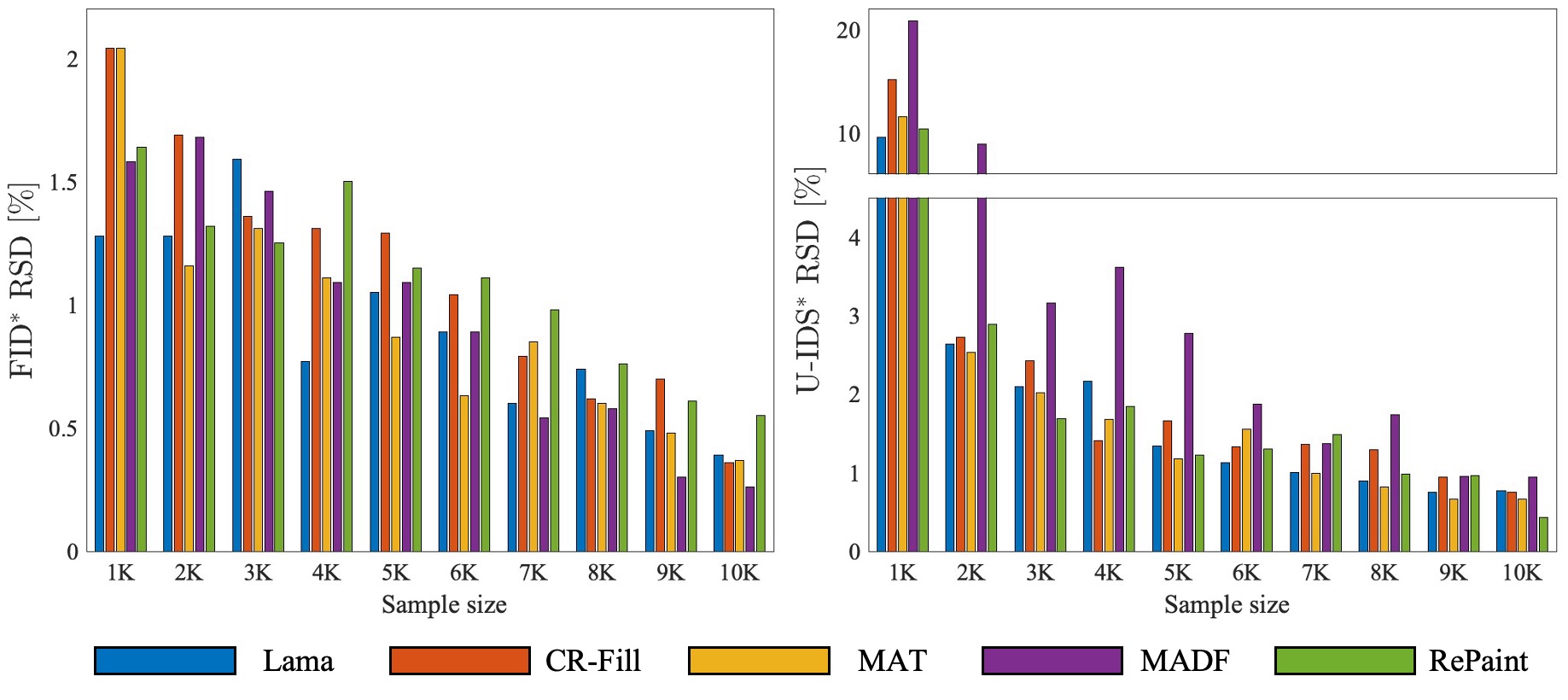}
  \vspace{-3mm}
  \caption{RSD values of FID$^*$ and U-IDS$^*$ obtained by setting the number of random samples differently. Results are obtained through 20 iterations. Lama \cite{suvorov2022resolution}, CR-Fill \cite{zeng2021cr}, MAT \cite{li2022mat}, MADF \cite{zhu2021image}, and RePaint \cite{lugmayr2022repaint}  are utilized to generate object removal results.}
    \label{fig:sample_num}
\vspace{-5mm}
\end{figure}

\textbf{Sample size for reliable evaluation.}
To examine the minimum sample size necessary for the proposed methods to reliably evaluate the performance of an object remover, we utilize the Relative Standard Deviation (RSD) which represents the deviation-to-mean ratio. We investigate the RSD of FID$^*$ and U-IDS$^*$ by varying the number of samples, which is presented in Fig. \ref{fig:sample_num}. To obtain query images, we randomly sample images from the COCO validation set containing \textit{person} class objects. Images from the COCO training set without \textit{person} class objects are worked as the comparison set. Input masks are generated using GT semantic segmentation masks for the target class objects. As shown in Fig. \ref{fig:sample_num}, the RSD values become less than 1\% when we use more than 7K samples for FID$^*$ and 9K samples for U-IDS$^*$. In other words, it is possible to reliably compare the performance of object removers without generating object removal GT of the COCO dataset when we use more than 9K object removal samples.

\vspace{-6mm}
\section{Conclusion}
We propose object remover performance evaluation methods that do not rely on an object removal ground truth. The proposed methods measure the performance using class-wise object removal results and the comparison set composed of images without target class objects. Experiments conducted on the COCO and self-acquired datasets demonstrate that the proposed methods can properly evaluate the object remover performance using images from various environments.

{\small
\bibliographystyle{IEEEtran}
\bibliography{bib_list}
}

\end{document}